\documentclass[twoside,11pt]{article}

\usepackage[preprint,nohyperref]{jmlr2e}

\usepackage{hyperref}
\usepackage{booktabs}
\usepackage{microtype}

\usepackage{amsmath}
\usepackage{mathtools}

\usepackage{booktabs}
\usepackage{longtable}
\usepackage{array}
\usepackage{multirow}
\usepackage{wrapfig}
\usepackage{float}
\usepackage{colortbl}
\usepackage{pdflscape}
\usepackage{tabu}
\usepackage{threeparttable}
\usepackage{threeparttablex}
\usepackage[normalem]{ulem}
\usepackage[utf8]{inputenc}
\usepackage{makecell}
\usepackage{xcolor}

\newcommand{\logit}{\operatorname{logit}}
\newcommand{\logitnormal}{\operatorname{LogitNormal}}
\newcommand{\sigmoid}{\operatorname{sigmoid}}
\newcommand{\bivariatenormal}{\operatorname{BN}}

\begin{document}

\title{Beyond Accuracy: How AI Metacognitive Sensitivity improves AI-assisted Decision Making}

\author{\name ZhaoBin Li \email zhaobin.li@uci.edu \\
       \addr Department of Cognitive Sciences \\
        University of California, Irvine
       \AND
       \name Mark Steyvers \email mark.steyvers@uci.edu \\
       \addr Department of Cognitive Sciences \\
        University of California, Irvine}
\editor{NA}

\maketitle

\begin{abstract}
In settings where human decision‐making relies on AI input, both the predictive accuracy of the AI system and the reliability of its confidence estimates influence decision quality. We highlight the role of AI metacognitive sensitivity---its ability to assign confidence scores that accurately distinguish correct from incorrect predictions---and introduce a theoretical framework for assessing the joint impact of AI's predictive accuracy and metacognitive sensitivity in hybrid decision‐making settings. Our analysis identifies conditions under which an AI with lower predictive accuracy but higher metacognitive sensitivity can enhance the overall accuracy of human decision making. Finally, a behavioral experiment confirms that greater AI metacognitive sensitivity improves human decision performance. Together, these findings underscore the importance of evaluating AI assistance not only by accuracy but also by metacognitive sensitivity, and of optimizing both to achieve superior decision outcomes.
\end{abstract}


\section{Introduction}
Humans are increasingly relying on artificial intelligence (AI) for decision support in domains such as forecasting \citep{benjamin2023hybrid}, clinical diagnosis, and judicial advisory services. The potential benefits of this collaboration stems from the complementary strengths and weaknesses of humans and AI. AI systems trained on large datasets can match or surpass human performance in narrow, specialized tasks, whereas humans excel at learning quickly from limited examples \citep{Lake2011-nm, Lee2015-ms}, adapting to novel situations \citep{Franklin2020-da, Wu2024-fo}, and juggling diverse objectives \citep{Goertzel2014-cc}. As a result, hybrid human-AI systems are expected to become increasingly common in everyday life.

\begin{figure}
    \centering
    \includegraphics[width=0.45\linewidth]{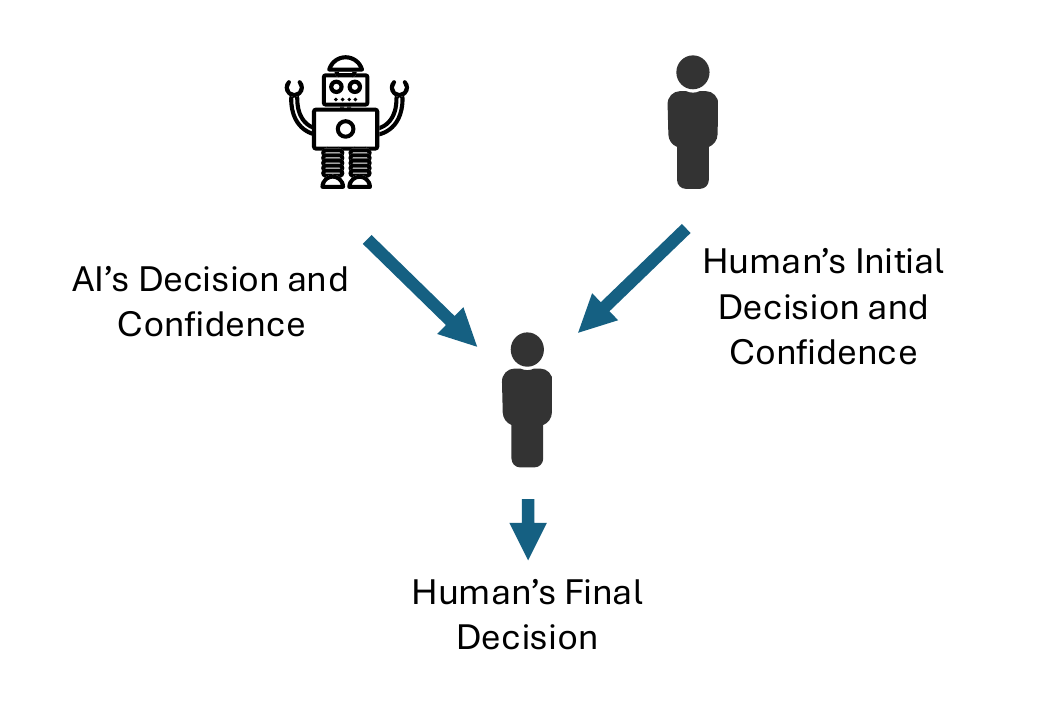}
    \hfill
    \includegraphics[width=0.53\linewidth]{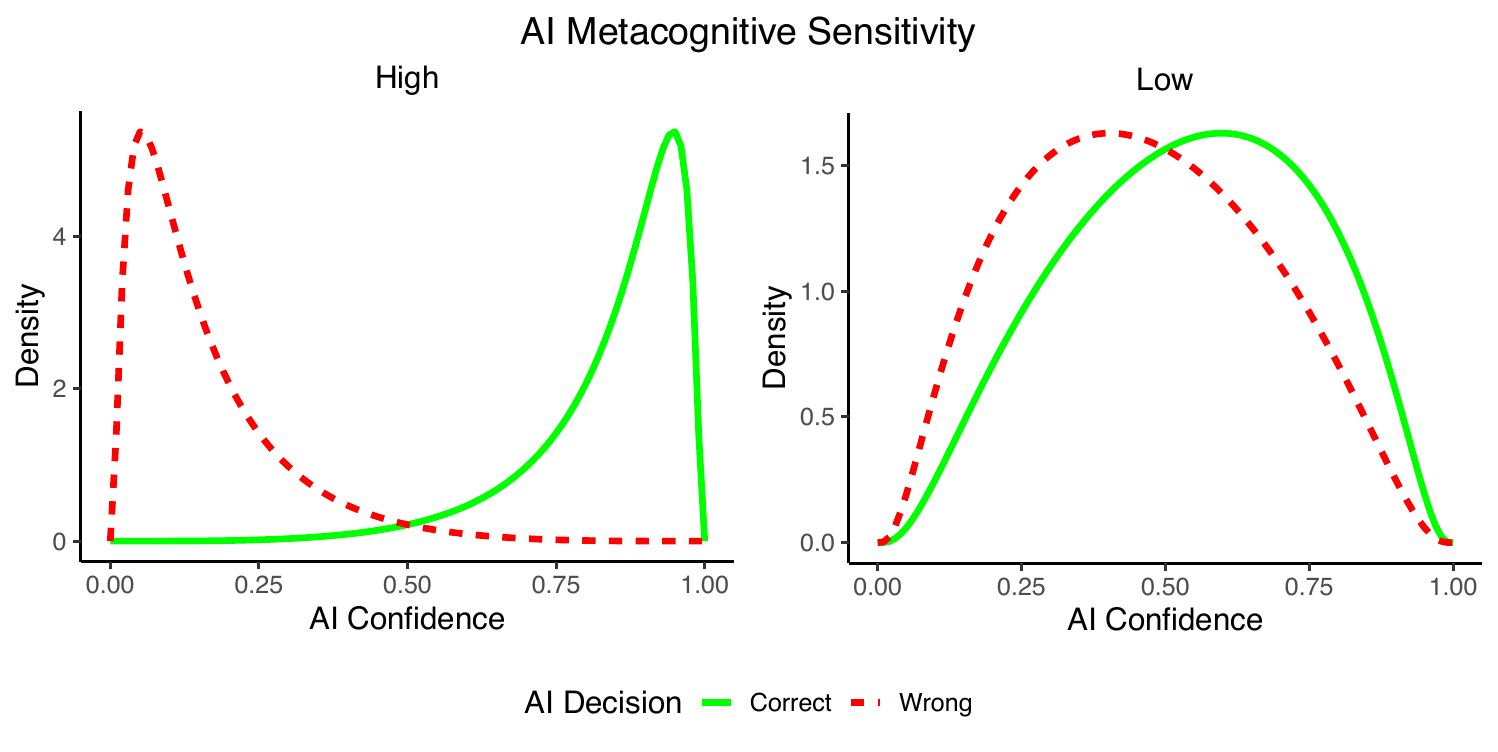}
    \caption{%
        \textbf{Left:} Illustration of the AI-assisted decision-making paradigm, in which a human integrates both the AI’s prediction and its associated confidence score to inform—and potentially improve—their final decision. \textbf{Right:} The AI’s metacognitive sensitivity measures the separation between its confidence distributions for correct and incorrect decisions. An AI with high metacognitive sensitivity is more likely to assign high confidence when it is correct and low confidence when it is wrong, compared to an AI with low metacognitive sensitivity. We show that an AI with high metacognitive sensitivity can assist humans in making better decisions.
    }
    \label{fig:assisted}
\end{figure}

In AI-assisted decision-making, an artificial agent provides a recommendation that a human decision-maker may accept or override (see Figure \ref{fig:assisted}). The challenge in this setup is to maximize expected utility, typically operationalized  as the accuracy of the human’s final judgment. Achieving this goal requires appropriate reliance: humans should accept the AI’s advice when it is likely to be correct and reject it otherwise. However, humans usually lack direct knowledge of whether a particular AI recommendation is right or wrong. While overall performance metrics such as accuracy on benchmark datasets provide broad guidance, they do not inform users about the reliability of individual predictions. Instead, humans must rely on instance-specific signals—most notably confidence scores, explanations, and their own mental model of the AI’s behavior—to decide when to trust a recommendation \citep{Steyvers2023-ve, Lee2025-jm}.

Among these signals, confidence scores are especially valuable. Just as humans intuitively estimate the certainty of their own judgments \citep{Peters2022-xl, Grimaldi2015-de}, many AI models output confidence estimates as a byproduct of their inference process. In classification settings, models such as logistic regression, naive Bayes, and neural networks with softmax outputs generate class probabilities directly. Ensemble methods like bagging or dropout approximate confidence via variability across model instances \citep{Breiman1996-vc,Gal2015-qk}. In large language models (LLMs), confidence estimation remains an open research area, with promising methods including token-wise likelihood analysis and sampling-based techniques \citep{Jiang2020-bv, Kadavath2022-fj}. When these confidence estimates track correctness reliably, they can guide users toward more accurate decisions by signaling when the AI’s prediction is likely to be right or wrong.

\subsection{Metacognitive Sensitivity versus Metacognitive Calibration}
The relationship between accuracy and confidence estimates can be characterized along two dimensions—\emph{metacognitive calibration} and \emph{metacognitive sensitivity}. Both play an important role in AI‐assisted decision‐making. Metacognitive calibration quantifies the correspondence between reported confidence and actual accuracy, measuring the degree to which the agent exhibits systematic overconfidence or underconfidence \citep{Fleming2014-wr}. Improved calibration in AI systems has been shown to enhance joint human–AI performance across a variety of tasks \citep{Ma2024-aq, Benz2023-vh, Zhang2020-yf}, although some studies report exceptions under specific conditions \citep{Vodrahalli2022-wh}.

In contrast, metacognitive sensitivity measures how well an agent’s confidence discriminates between correct and incorrect predictions \citep{Fleming2014-wr}. In the forecasting literature, this is also referred to as discrimination \citep{mandel2014accuracy} or resolution \citep{Murphy1973-ap, Siegert2017-fs}. Recent work by \citet{Steyvers2025-yc} demonstrates that LLMs  with high metacognitive sensitivity—communicating their confidence effectively through verbal explanations—enable human collaborators to more accurately identify which AI responses are likely to be correct.

Although related, calibration and sensitivity capture distinct aspects of metacognitive performance. Calibration reflects how closely confidence scores align with empirical accuracy and can often be improved post hoc via recalibration techniques such as Platt scaling \citep{Platt1999-bn}. Sensitivity, however, reflects the model’s intrinsic ability to rank correct predictions above incorrect ones and typically requires architectural or training modifications to enhance. 



\subsection{The Need for a Theoretical Framework}
Despite growing interest in how AI confidence interacts with human decision‐making, prior research has not explicitly isolated the role of AI metacognitive sensitivity nor examined its trade‐offs with base accuracy. In this research, we address this gap by proposing a theoretical model grounded in signal detection theory \citep{Maniscalco2012-xx}. This framework jointly models an AI’s accuracy and metacognitive sensitivity and analyzes their effect on the performance of a human decision‐maker assisted by the AI. 


In this framework, we derive an analytic expression for the human’s accuracy after receiving AI assistance as a function of both AI accuracy and metacognitive sensitivity. Analytical results under plausible assumptions reveal conditions in which an AI system with lower accuracy but higher metacognitive sensitivity may outperform a more accurate but less metacognitively sensitive AI in supporting human decision-making. We refer to this as an \textit{inversion scenario}. 

We validate these theoretical predictions in a behavioral experiment in which human participants make perceptual judgments with AI assistance that vary systematically in both accuracy and sensitivity. The experimental results confirm the theoretical model,  demonstrate the occurrence of inversion scenarios, and underscore the necessity of balancing both accuracy and metacognitive sensitivity to optimize AI‐assisted decision‐making.

\section{Problem Setup}

We study a decision-making problem where a human decision-maker faces a multi-class classification task. For each instance, the human must choose whether to rely on their own judgment or on the recommendation of an AI model. Let $Y \in \{1, \dotsc, K\}$ denote the true class label for a given instance, where $K \geq 2$ is the total number of possible labels. The human produces a predicted label $\hat{Y}_h \in \{1, \dotsc, K\}$, and the AI model produces a predicted label $\hat{Y}_m \in \{1, \dotsc, K\}$. Along with these predictions, both the human and the AI provide associated confidence scores, denoted by $c_h$ and $c_m$, respectively, where $c_h, c_m \in [0,1]$. The correctness of each prediction is captured by a binary outcome variable: $y_h = 1$ if $\hat{Y}_h = Y$ (the human prediction is correct), and $y_h = 0$ otherwise, and $y_m = 1$ if $\hat{Y}_m = Y$ (the AI prediction is correct), and $y_m = 0$ otherwise

At each decision point, the human selects an action $A \in \{H, M\}$ defined as
\begin{equation}
A =
\begin{cases}
H, & \text{rely on own prediction } \hat{Y}_h, \\
M, & \text{rely on AI prediction } \hat{Y}_m.
\end{cases}
\end{equation}

The human’s objective is to select the action that maximizes their expected utility. The realized utility depends on whether the selected prediction matches the true label. Formally, the utility function is defined as:
\begin{equation}
U(A) =
\begin{cases}
1, & \text{if the selected prediction } \hat{Y}_A = Y, \\
0, & \text{otherwise}.
\end{cases}
\end{equation}

Thus, a correct classification by the chosen agent yields a utility of 1, while an incorrect classification yields a utility of 0. Consequently, under this utility structure, the human’s decision problem corresponds to maximizing their overall classification accuracy. Although we primarily focus on this symmetric utility setting, the model can be extended to allow asymmetric payoffs, where correct and incorrect classifications carry different utilities depending on the application.

Importantly, the confidence scores $c_h$ and $c_m$ are not assumed to be perfect indicators of correctness. These scores may be noisy, biased, or otherwise imperfect reflections of the probability that a given prediction is correct. Therefore, how the human interprets and combines the available confidence information plays a critical role in achieving high decision quality.

A central concept in this analysis is \emph{metacognitive sensitivity}, which refers to the extent to which an agent’s confidence distinguishes between correct and incorrect predictions. An agent with high metacognitive sensitivity tends to assign higher confidence scores to correct predictions and lower confidence scores to incorrect ones. In this work, we focus specifically on the metacognitive sensitivity of the AI model and study how variations in this property influence the overall accuracy of the human-AI decision-making system.


\section{A Signal Detection Framework to Analyze Accuracy and Metacognitive Sensitivity}

To analyze the interplay between accuracy and metacognitive sensitivity in AI models, we adopt a signal detection theory framework to model the generation of confidence scores \citep{Galvin2003-tb,steyvers2014evaluating}. This framework provides a principled way to simulate and manipulate both the accuracy and metacognitive sensitivity of the AI advice. Within this framework, we derive analytic expressions for the decision accuracy of the human assisted by AI advice.

Our analysis highlights that metacognitive sensitivity can be as important as, or even more important than, the AI’s baseline accuracy in certain settings. Specifically, an AI model with lower overall accuracy but higher metacognitive sensitivity may enable the human to achieve better outcomes than a more accurate but less metacognitively sensitive AI. Thus, enhancing the AI's ability to discriminate between correct and incorrect predictions can be an effective way to improve human-AI team performance.


\subsection{AI Confidence Distribution}
Let \( y_m \) be a Bernoulli random variable representing the correctness of a prediction, with latent variable \( \theta_m \) representing the probability of a correct model prediction, i.e.\ \( y_m = 1 \): 
\begin{equation}
y_m \sim \mathrm{Bernoulli}(\theta_m)
\end{equation}
The AI model generates a confidence score \( c_m \) conditioned on \( y_m \). We assume the confidence distributions for correct and incorrect predictions follow logit-normal distributions:
\begin{equation}
c_m \mid y_m \sim 
\begin{cases} 
\logitnormal(\mu_0, \sigma^2) & \text{if } y_m = 0, \\
\logitnormal(\mu_1, \sigma^2) & \text{if } y_m = 1.
\end{cases}
\end{equation}
where \( \mu_1 > \mu_0 \) reflects the model's ability to assign higher confidence to correct predictions. The marginal confidence distribution is given by: \( p(c_m) = p(y_m = 1) p(c_m \mid y_m=1) + p(y_m = 0) p(c_m \mid y_m=0) \) where \( p(y_m = 1) \) is the model accuracy \( \theta_m \) and \( p(y_m = 0) = 1 - \theta_m \). 

\subsection{Metacognitive Sensitivity}
Metacognitive sensitivity reflects the model’s ability to assign higher confidence scores to correct predictions than to incorrect ones. We can quantify this using two related metrics. First, Cohen’s \( d \)  measures the standardized mean difference between the confidence distributions for correct and incorrect predictions:
\begin{equation}
d = \frac{\mu_1 - \mu_0}{\sigma}.
\end{equation}
As an alternative metric, we can also convert \( d \) to the area under the curve (AUC) for ease of interpretation. The AUC in our context is the probability that a randomly selected correct prediction will have a higher confidence score than an incorrect one. For normal distributions, this type of AUC can be directly derived from \( d \) \citep{Kraemer2014-or}:
\begin{equation}
\text{AUC} = \Phi\left(\frac{d}{\sqrt{2}}\right),
\end{equation}
where \( \Phi \) is the standard normal cumulative distribution function. The AUC metric for metacognitive sensitivity is \emph{not} the same as the typical ROC AUC used in machine learning. The ROC AUC assesses the discrimination for positive and negative labels, whereas in the signal detection framework we apply here, the AUC assesses the discrimination for correct and incorrect decisions.

\begin{figure}[tb]
    \centering
    \includegraphics[width=0.45\linewidth]{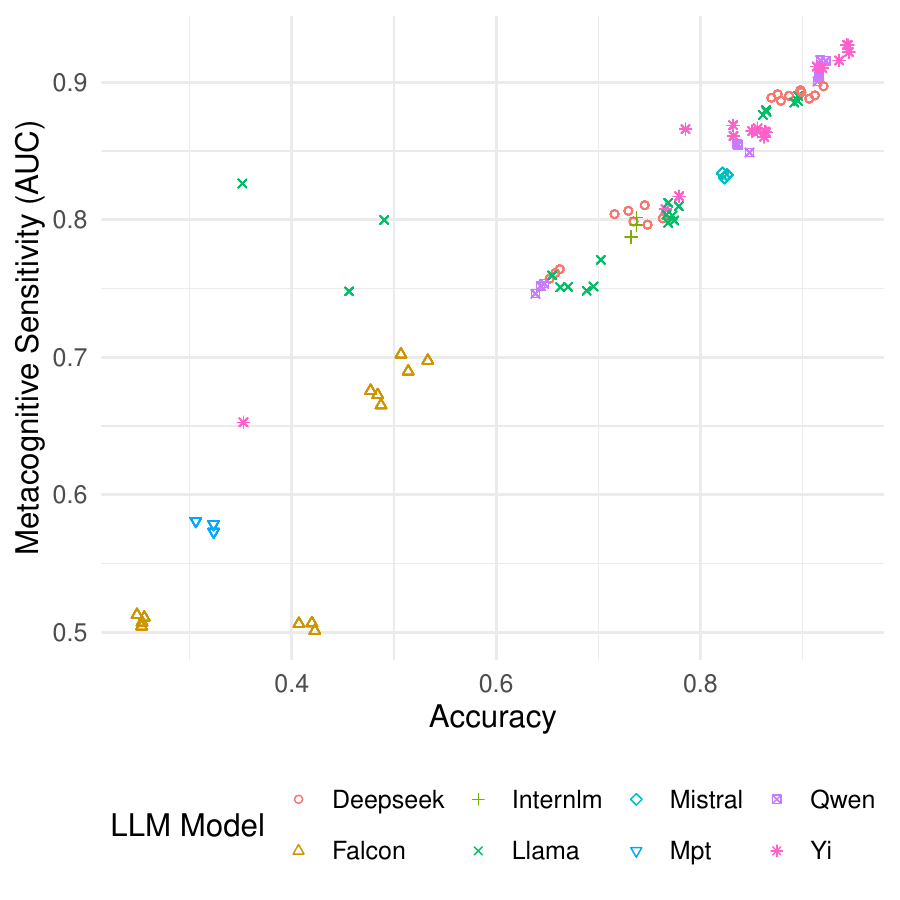}
    \hfill
    \includegraphics[width=0.45\linewidth]{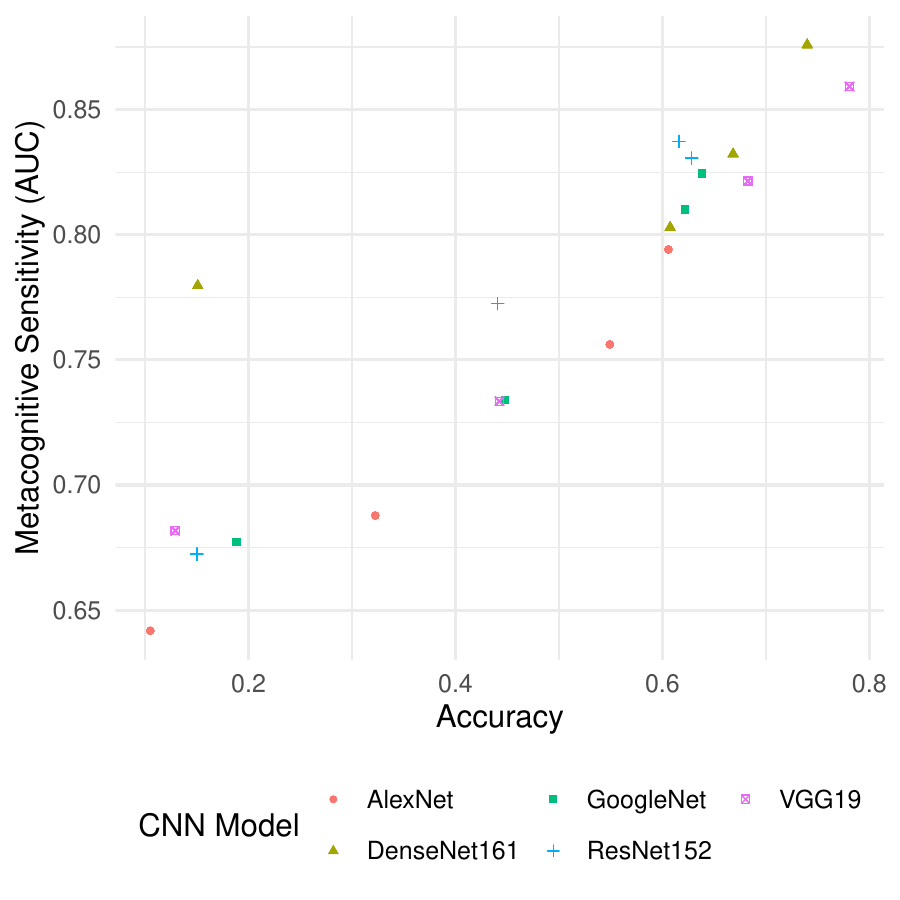}
    \caption{Relationship between metacognitive sensitivity (measured using AUC) and accuracy for LLMs (left) and CNNs (right) across several benchmark tasks and model architectures. The results reveal substantial variations in metacognitive sensitivity within the LLM models.}
    \label{fig:auc_vs_acc}
\end{figure}

To illustrate how an AI model’s metacognitive sensitivity varies in real-world settings—and how this relates to the model’s accuracy—we plot these metrics for two types of AI models: convolutional neural networks (CNNs) for computer-vision tasks and large language models (LLMs) for natural-language-processing benchmarks (Fig.~\ref{fig:auc_vs_acc}). The results reveal a broad spectrum of metacognitive sensitivities across models, particularly among LLMs, and show that sensitivity is not perfectly correlated with accuracy. Further experimental details are provided in Appendix~\ref{app:addmd}.

\subsection{Expected utility with constant human confidence}\label{sec:con}
When a human decision-maker combines their judgment with AI predictions, the goal is to maximize the combined decision accuracy. For a given input, the human decides whether to accept the AI’s prediction based on the AI’s confidence score \( c_m \) and their own confidence score \( c_h \). We first consider the case when human confidence is constant across all problems. For simplicity, we also assume perfect calibration for the human: when a human assigns a confidence \( c_h \), their accuracy is also $c_h$ on average. 

Using Bayes' rule, the model calibration function that relates model accuracy to model confidence is 
\[ 
p(y_m=1 \mid c_m) = \frac{p(y_m=1) p(c_m \mid y_m=1)}{p(c_m)}. 
\] 
Since \(\mu_1 > \mu_0\) for the model confidence distributions, the probability of a correct model prediction conditional on confidence, \( p(y_m=1 \mid c_m) \), increases monotonically with \(c_m\). Hence, as the model confidence increases up to a certain point, which we shall call the switch-point \( c_m = c^* \), the human accuracy is equal to AI accuracy:
\begin{equation}
c_h = p(y_m=1 \mid c^*).
\end{equation}

Hence, the optimal policy is for the human decision-maker to go with their own decision if the model confidence is less than the switch-point confidence (\(c_m < c^*\)),  because the human is more accurate. If not, the human decision-maker should go with the model decision because the model is more accurate:

\begin{equation}
A = 
\begin{cases} 
H & \text{if } c_m < c^*, \\
M & \text{if } c_m \geq c^*.
\end{cases}
\end{equation}

The probability of a correct final human prediction (after taking into account the AI advice) for a particular level of human confidence \( c_h \) is given by:
\begin{equation}
p(y_{m,\,h} = 1 \mid c_h) = 
\begin{cases} 
c_h & \text{if } A=H, \\
p(y_m=1 \mid c_m) & \text{if } A=M.
\end{cases}
\end{equation}

Thus, the combined accuracy can be expressed as follows:

\begin{multline}
p(y_{m,\,h} = 1 \mid c_h) = \int_0^{c^*} c_h p(c_m) \, dc_m \\ 
+ \int_{c^*}^1 p(y_m=1 \mid c_m) p(c_m) \, dc_m,    
\end{multline}
To find \(c^*\), we invert \(f\) by setting:
\begin{equation}
c_h = f(c^*).
\end{equation}
Solving this yields (Appendix \ref{sec:appdxa}):

\begin{align}
c^* = \sigmoid&\Bigl(\frac{\sigma^2[\logit(c_h) - \logit(\theta_m)]}{\mu_1 - \mu_0}  \notag\\ 
& \quad + \frac{\mu_1 + \mu_0}{2}\Bigr),
\end{align}

Using properties of the logit-normal distribution and expanding $c^*$, the combined accuracy and therefore the expected utility becomes (Appendix  \ref{sec:appdxb}):


\begin{align} \label{eq:ycombinedconstant}
\mathbb{E}[U(A)\mid c_h] &= p(y_{m,\,h} = 1 \mid c_h) \notag\\ 
&= c_h\Bigl[\theta_m\Phi(r) \notag\\ 
& \quad + (1-\theta_m)\Phi(r+d)\Bigr] \notag\\ 
& \quad + \theta_m \Bigl[1 - \Phi(r)\Bigr].  
\end{align}
where
\begin{equation}
r = \frac{\logit(c_h) - \logit(\theta_m)}{d} - \frac d 2
\end{equation}
Note that \( p(y_{m,\,h} = 1 \mid c_h) \) depends only on: the human confidence $c_h$, the model accuracy  $\theta_m$ and the model's metacognitive sensitivity $d$.

\begin{figure}[tb]
    \centering
    \includegraphics[width=0.7\linewidth]{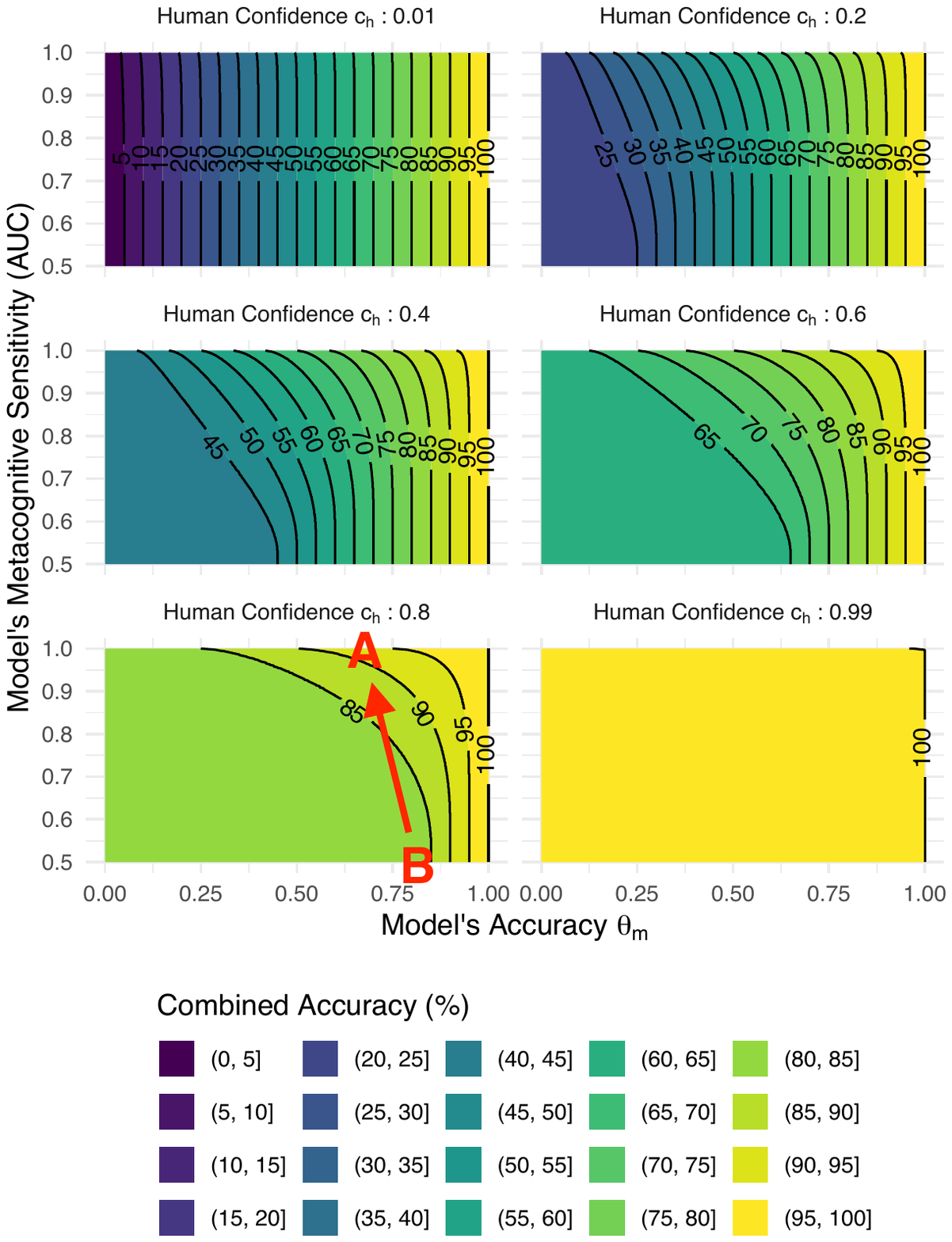}
    \caption{Predicted combined human-AI accuracy as a function of model accuracy ($\theta_m$) and metacognitive sensitivity (assessed by AUC) shown across panels representing fixed levels of human confidence $c_h$ (panels). The red arrow shows a scenario where a model with lower accuracy but  higher sensitivity (Model A) outperforms a more accurate but less sensitive model (Model B) when combined with a human.}
    \label{fig:constant}
\end{figure}

Figure \ref{fig:constant} shows the predicted combined accuracy as a function of model accuracy and model discrimination (measured by AUC) across different levels of (fixed) human confidence. From the resulting contours, we can find examples where model A has lower accuracy but higher metacognitive sensitivity than model B, and achieves higher combined accuracy when paired with a human.

The upper-left panel of Figure \ref{fig:constant} provides an important boundary condition for our analysis: when the human decision-maker is performing at chance (i.e., $c_h$ = 0.01), there are no inversion scenarios. In this regime, the human should always defer to the AI’s recommendation. As a result, the final decision accuracy simply reflects the AI's own accuracy, irrespective of its metacognitive sensitivity. This leads to a monotonic relationship where the best combined performance is achieved by the AI with the highest predictive accuracy. Since the human contributes no useful information, differences in metacognitive sensitivity (AUC) do not affect outcomes in this region of the parameter space.

\subsection{Expected utility with variable human confidence}\label{sec:var}
We now consider the more general case when human confidence varies across problems. Specifically, we will assume that human confidence follows a logit-normal distribution:
\begin{equation}
c_h \sim \logitnormal(\mu_h, \sigma_h).
\end{equation}
As before, we assume that the human decision-maker is perfectly calibrated (i.e., the confidence \( c_h \) directly reflects the human's probability of correctness).  

To derive the predicted combined accuracy \( p(y_{m,\,h} = 1) \) and therefore the expected utility $\mathbb{E}[U(A)]$, we integrate the accuracy conditional on human confidence \( p(y_{m,\,h} = 1 \mid c_h) \) over \( c_h \) (Appendix \ref{sec:appdxc}):

\begin{align} \label{eq:ycombinedvariable}
\mathbb{E}[U(A)] &= p(y_{m,\,h} = 1) \notag\\
&=\int_0^1 p(c_h) p(y_{m,\,h} = 1 \mid c_h) \, dc_h \notag\\
&\approx \theta_m \bivariatenormal\left(\frac{a}{\sqrt{1+b^2}}, \frac{s}{\sqrt{1+t^2}}, \right. \notag\\
&\qquad\qquad \quad \left. \rho=\frac{b t}{\sqrt{1+b^2} \sqrt{1+t^2}}\right) \notag\\
& + (1-\theta_m) \bivariatenormal\left(\frac{a + d}{\sqrt{1+b^2}}, \frac{s}{\sqrt{1+t^2}}, \right. \notag\\
&\qquad\qquad \quad \left. \rho=\frac{b t}{\sqrt{1+b^2} \sqrt{1+t^2}}\right) \notag\\
& + \theta_m\left[1 - \Phi\left(\frac{a}{\sqrt{1+b^2}}\right)\right]
\end{align}
where

\begin{gather}
a=\frac{\mu_h - \logit(\theta_m)}d - \frac d 2, \quad b=\frac {\sigma_h} d \notag\\
s=\lambda \mu_h, \quad t=\lambda \sigma_h
\end{gather}

The operator $\bivariatenormal$ is the Bivariate-Normal CDF and $\lambda=\sqrt{\frac{\pi}8}$ is the constant for probit approximation. We tested various parameters and the approximation is highly accurate, within 1\% difference of numerical integration.

The overall combined accuracy depends only on: the mean $\mu_h$ and variance $\sigma_h$ of the human confidence distribution, the model accuracy $\theta_m$, and the model's metacognitive sensitivity $d$.

\begin{figure}[tb]
    \centering
    \includegraphics[width=0.8\linewidth]{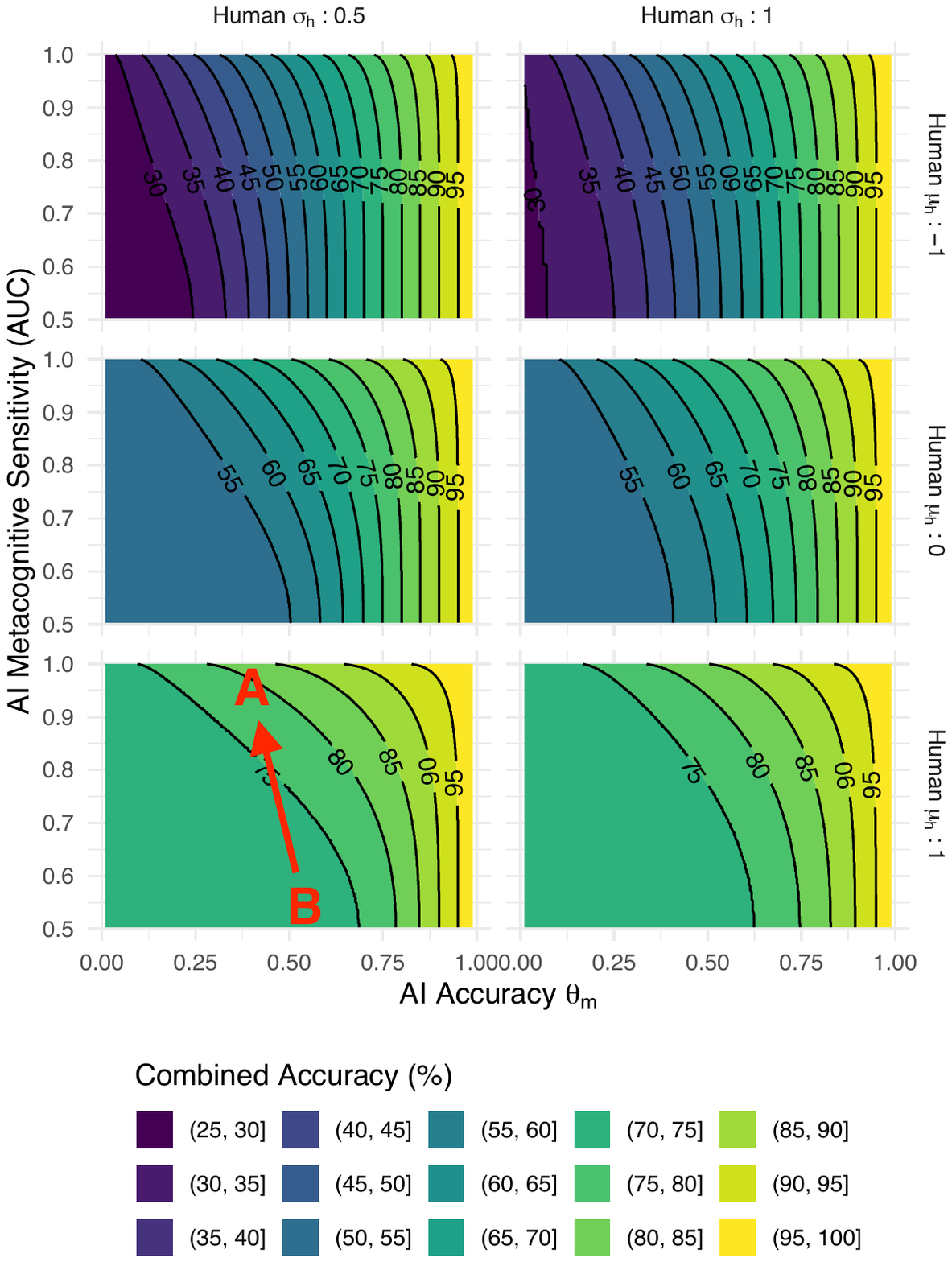}
    \caption{Predicted combined human-AI accuracy as a function of AI model accuracy ($\theta_m$) and metacognitive sensitivity (assessed by AUC) across panels showing different distributions of human confidence, characterized by means $\mu$ and standard deviation $\sigma$. The red arrow shows a  scenario in which a model with lower accuracy but  higher metacognitive sensitivity (Model A) achieves superior combined accuracy compared to a more accurate but less sensitive model (Model B).}
    \label{fig:variable}
\end{figure}

In Figure \ref{fig:variable}, we graph the combined accuracy based on these four variables. The pattern of contours is similar to the graph of constant human confidence. Once again, we can point to an example where model A has lower accuracy but higher metacognitive sensitivity than model B, yet achieves higher combined accuracy when paired with a human.

\section{Behavioral Experiment}
To investigate whether an AI’s metacognitive sensitivity influences human decision-making, we conducted an online behavioral experiment. Building on prior research on collaborative decision-making between humans and AI \citep{Tejeda2022-wv,liang2022adapting}, we used a perceptual task in which participants viewed noisy animations containing dots of various colors. Their objective was to identify the color with the highest number of dots. Participants were provided with both predictions and confidence scores from an AI assistant and were instructed to use this information to improve their own responses. The AI assistance varied across participants in both accuracy and metacognitive sensitivity.

According to our theoretical model, we hypothesized that, when AI accuracy is held constant, greater metacognitive sensitivity would lead to better combined human-AI performance. We also considered inversion scenarios in which an AI with lower overall accuracy but higher metacognitive sensitivity might result in superior collaborative outcomes compared to an AI with higher accuracy but lower metacognitive sensitivity.

\subsection{Participants}
We recruited 110 participants online via Prolific. All participants were based in the United States and ranged in age from 20 to 65, with 52\% men and 48\% women. Each participant was compensated \$6 for completing the study, which lasted approximately 30 minutes. The study protocol was approved by the university's Institutional Review Board (IRB). To ensure participants understood the experimental instructions, we included an onboarding process using the Intro.js interface tutorial system. 

\subsection{Materials}
\begin{figure}[ht]
    \centering
    \includegraphics[width=\linewidth]{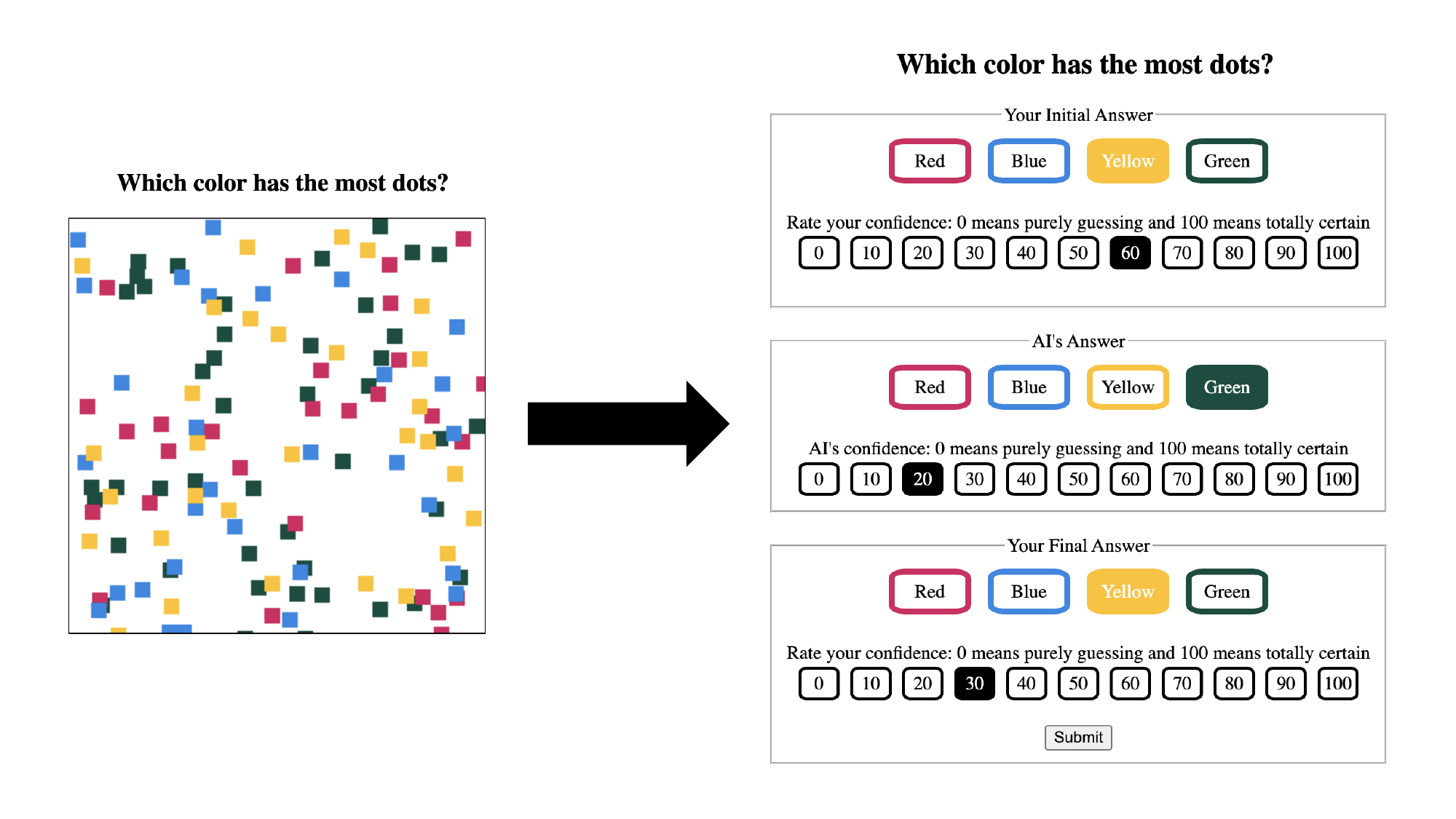}
    \caption{Example of the visual display and user interface presented during the behavioral experiment. Participants viewed the 400×1200 pixel canvas through a 400×400 pixel sliding window, creating a smooth scrolling animation that revealed the dot field over a 3-second period.}
    \label{fig:userinterface}
\end{figure}

Each trial consisted of a 3-second animation in which colored dots moved downward on the screen. Participants were asked to identify the color that appeared most frequently (see Figure \ref{fig:userinterface} for an example). The four possible colors were red, blue, yellow, and green. For each trial, one color was randomly designated as the majority color, appearing 150 times, while the remaining three colors served as  minority colors, each appearing 100 times. Dots were displayed as 15$\times$15 pixel squares, randomly distributed across a 400$\times$1200 pixel canvas. Participants viewed the canvas through a 400$\times$400 pixel sliding window, which created a smooth scrolling animation over the 3-second period.

\subsection{Procedure}
We used a between-subjects design in which each participant was randomly assigned to one of five AI assistants that varied in accuracy and metacognitive sensitivity (see top two rows of Table~\ref{tab:ai}). For four of the AI conditions, AI accuracy was set to 0.66, a level high enough to offer useful guidance but not so high that participants would follow the AI uncritically. These assistants varied in metacognitive sensitivity, with AUC values at 0.50, 0.76, 0.89, and 0.99, covering a range from chance-level (AUC=0.5) to near optimal (AUC=0.99) metacognitive sensitivity.

To test whether an AI with lower accuracy but higher metacognitive sensitivity could still lead to better joint performance---as predicted by our theoretical model---we included a fifth assistant with 0.55 accuracy and an AUC of 0.99. This setting represents a realistic inversion scenario, since pilot studies showed that 0.55 falls within the typical range of unaided human performance.

On each trial, participants first viewed the animation and made an initial judgment about the majority color, followed by a confidence rating on a scale from 0 (pure guess) to 100 (total certainty) (Figure \ref{fig:userinterface}). They then received the AI’s prediction along with its confidence rating. Based on this information, participants could revise both their decision and their confidence. The correct answer was revealed at the end of each trial. Each participant completed 100 trials. Their final utility was based on the accuracy of their responses after the opportunity to revise.

\section{Results}
A total of 110 participants were randomly assigned to the five AI conditions (with 21–23 participants per condition). Before receiving AI advice, participants had a mean accuracy of 0.55 (SD = 0.12). After incorporating  the AI's input, mean accuracy increased to 0.73 (SD = 0.089), resulting in an average gain of 0.18 (SD = 0.10). A summary of the descriptive statistics is shown in Table \ref{tab:ai}.

\begin{table}
\caption{Results of the behavioral experiment across the five AI conditions. Standard deviations are shown in parentheses.}

\centering
\begin{tabular}[t]{llllll}
\toprule
\multicolumn{1}{c}{ } & \multicolumn{5}{c}{AI Conditions} \\
\cmidrule(l{3pt}r{3pt}){2-6}
Metric & A & B & C & D & E\\
\midrule
\addlinespace[0.3em]
\multicolumn{6}{l}{\textbf{AI}}\\
\hspace{1em}Accuracy & 0.66 & 0.66 & 0.66 & 0.66 & 0.55\\
\hspace{1em}Metacognitive Sensitivity (AUC) & 0.5 & 0.76 & 0.89 & 0.99 & 0.99\\
\addlinespace[0.3em]
\multicolumn{6}{l}{\textbf{Human}}\\
\hspace{1em}Accuracy before AI advice & 0.57 (0.11) & 0.55 (0.12) & 0.53 (0.14) & 0.53 (0.13) & 0.57 (0.13)\\
\hspace{1em}Accuracy after AI advice & 0.67 (0.06) & 0.69 (0.09) & 0.73 (0.07) & 0.78 (0.1) & 0.78 (0.07)\\
\bottomrule
\end{tabular}

\label{tab:ai}
\end{table}

\subsection{Effect of AI Metacognitive Sensitivity}
We first tested whether, when holding AI accuracy constant at 0.66, higher AI metacognitive sensitivity (as measured by AUC) would lead to improved human–AI performance. A multilevel logistic regression was conducted to predict trial‐level accuracy after advice (correct vs.\ incorrect). The predictors included (1) the correctness of the participant’s initial response and (2) the AI’s AUC, with random intercepts and slopes for each participant. The analysis revealed a significant positive effect of AI metacognitive sensitivity, $\beta = 1.62$, SE = 0.24, $z = 6.87$, $p < .001$. Thus, participants paired with more metacognitively sensitive AIs achieved higher accuracy after receiving advice.

\subsection{AI Metacognitive Sensitivity Improves Human–AI Complementarity}
Next, we examined complementarity, defined as the extent to which combined human–AI accuracy exceeded the better of either the human’s unaided accuracy or the AI’s own accuracy. For each AI (all with 0.66 accuracy), we computed two metrics across participants.

First, we measured the proportion of participants for whom final joint accuracy exceeded both initial human accuracy and AI accuracy. A logistic regression predicting this incidence of complementarity from AI AUC yielded $\beta = 3.95$, SE = 1.32, $z = 3.00$, $p = .003$, indicating that higher AUC significantly increased the likelihood of achieving complementarity.

Second, we calculated the mean difference between joint accuracy and the higher of initial human accuracy or AI accuracy. A linear regression predicting this magnitude of complementarity from AI AUC yielded $\beta = 0.20$, SE = 0.04, $t = 4.70$, $p < .001$, showing that greater metacognitive sensitivity produced larger gains beyond the best single agent.

\subsection{Trading Accuracy for Metacognitive Sensitivity}
Finally, we tested whether a fifth AI with lower task accuracy but very high metacognitive sensitivity (0.55 accuracy, AUC = 0.99) could outperform higher‐accuracy (0.66) AIs with lower sensitivity. We compared final accuracy across the fifth AI and three of the 0.66‐accuracy AIs (AUCs = 0.50, 0.76, 0.89) using a multilevel logistic regression predicting trial‐level accuracy from initial human correctness and AI condition (with by‐participant random effects), followed by Dunnett’s post‐hoc tests with fifth AI as the reference group. Relative to the fifth AI, the higher‐accuracy but lower‐sensitivity AIs showed significantly lower odds of improving human decisions (odds ratios = 0.49, 0.56, and 0.74 for AUCs 0.50, 0.76, and 0.89, respectively; all $p < .05$). These results confirm that metacognitive sensitivity can outweigh task accuracy in driving effective human–AI collaboration.

\section{Discussion}

Both the theoretical and empirical results demonstrate the importance of an AI’s metacognitive sensitivity in AI-assisted decision-making: the higher the AI’s metacognitive sensitivity, the greater the expected utility obtained by the human decision-maker, as measured by final accuracy. The theoretical and empirical findings also confirm the existence of inversion scenarios, in which an AI with lower accuracy but higher metacognitive sensitivity can provide greater utility to the human decision-maker than an AI with higher accuracy but lower metacognitive sensitivity. Furthermore, when an AI with higher metacognitive sensitivity assists a human decision-maker, it is more likely to lead to human–AI complementarity, where their combined utility surpasses what either can achieve on their own. Together, these results highlight how metacognitive sensitivity helps human decision-makers discriminate when the AI’s assistance is correct or incorrect and thereby use its recommendations appropriately.

One limitation of our theoretical framework is the assumption that the human acts as an ideal observer—well calibrated, fully informed of the AI’s confidence–calibration curve, and capable of integrating all available information rationally to reach the optimal decision. Although humans are not ideal observers, the two main trends predicted by our framework still hold: (1) improvements in an AI’s metacognitive sensitivity increase the human decision-maker’s expected utility; and (2) inversion scenarios can arise in which superior metacognitive sensitivity compensates for lower AI accuracy. Our normative framework thus provides an upper bound on human expected utility. Future research should examine the cognitive processes by which humans interpret and integrate AI confidence signals and apply these insights to real-world decision-making contexts.

There are two natural extensions to the problem setup that could further explore the role of AI’s metacognitive sensitivity in AI-assisted decision-making. First, we can explore scenarios involving multiple humans and multiple AIs. When a single AI assists a human, we showed that if the human’s accuracy is very low, the AI’s metacognitive sensitivity matters little, because the human should always follow the AI’s decision. However, when the decision-maker can consult multiple humans or AIs, using a metacognitively sensitive AI will likely increase expected utility— the decision-maker can rely on the AI’s advice when it is confident and consult another human or AI when it is not.

Another possible extension is to broaden the definition of expected utility beyond the final accuracy of the human decision-maker to include aspects such as time and computational cost. In these scenarios, an AI’s metacognitive sensitivity can become even more important. For instance, large reasoning LLMs are usually more accurate but also require more compute power and are slower to provide an answer. If a smaller LLM with high metacognitive sensitivity is available, the decision-maker can consult it first and then defer to the larger LLM if the smaller model expresses uncertainty, thereby increasing expected utility. These ideas relate to the learning-to-defer literature with multiple experts \citep{Madras2017-hw,Keswani2021-vb}, illustrating the broader implications of AI’s metacognitive sensitivity.


Improving an AI’s metacognitive sensitivity may also mitigate human under‐ and over‐reliance—a persistent issue in human–AI collaboration \citep{Zhang2020-gi,Gaube2025-fo}. When a metacognitively sensitive AI’s confidence is substantially higher when it is correct than when it is wrong, humans can use those signals to decide when to accept or reject the AI’s recommendations \citep{Steyvers2025-yc}. 

These implications underscore the importance of understanding and improving AI metacognitive sensitivity, especially in large language models (LLMs). People rely on LLMs for advice on critical decisions—including legal, financial, and personal matters \citep{Imagining-the-Digital-Future-Center2025-xb}—and may over-rely on them when they do not know the model’s limitations. Indeed, the OWASP Top 10---a report by global security experts---identifies AI over-reliance as a major vulnerability in LLM deployments \citep{Wilson2023-vi}. A metacognitively sensitive LLM could mitigate this problem by accurately signaling its uncertainty. However, methods to reliably measure and enhance an LLM’s metacognitive sensitivity remain an active area of research, and developing such methods will become increasingly important as LLMs are integrated into everyday decision-making.

All of these points connect with the broader literature on human–AI interaction and the need to jointly optimize human–AI team performance \citep{Bansal2019-kl,Rechkemmer2022-ah}. As evidenced by AI leaderboards, benchmarks, and research papers, the AI community has traditionally emphasized accuracy as the primary—if not sole—metric for evaluating and comparing AI systems. While accuracy is paramount when AI systems operate in isolation, it is equally critical to account for collaborative factors—such as confidence signaling and the human integration of those signals—when AI is deployed as an aid to human decision-making. Doing so will maximize overall expected utility.

%
%
%
\section*{Appendices}
\vspace{1em}
\appendix
\section{Derivation of Switch point $c^*$}
\label{sec:appdxa}

To note, let $X \sim \logitnormal(\mu, \sigma)$. Then its CDF is given by $F_X(x) = \Phi\left(\frac{\logit(x) - \mu}{\sigma}\right)$ and its PDF is given by $f_X(x) = \frac1{\sigma x (1-x)}\phi\left(\frac{\logit(x) - \mu}{\sigma}\right)$. $\Phi$ and $\phi$ are the standard normal CDF and PDF.

To get $c^*$, we inverse $f$:

$$c_h = f(c^*)$$

$$c_h = \theta_m p(c^* | y = 1) / [\theta_m p(c^* | y = 1) + (1-\theta_m) p_{y_0}(c^*)]$$

$$c_h = \theta_m e^{-\frac{(\logit(c^*) -  \mu_1)^2}{2\sigma^2}}/[\theta_m e^{-\frac{(\logit(c^*) -  \mu_1)^2}{2\sigma^2}} + (1-\theta_m) e^{-\frac{(\logit(c^*) -  \mu_0)^2}{2\sigma^2}}]$$

$$c_h \theta_m e^{-\frac{(\logit(c^*) -  \mu_1)^2}{2\sigma^2}} + c_h (1-\theta_m) e^{-\frac{(\logit(c^*) -  \mu_0)^2}{2\sigma^2}} = \theta_m e^{-\frac{(\logit(c^*) -  \mu_1)^2}{2\sigma^2}}$$

$$c_h (1-\theta_m) e^{-\frac{(\logit(c^*) -  \mu_0)^2}{2\sigma^2}} = (1 - c_h) \theta_m e^{-\frac{(\logit(c^*) -  \mu_1)^2}{2\sigma^2}}$$

$$\log{c_h} + \log{(1-\theta_m)} - \frac{(\logit(c^*) -  \mu_0)^2}{2\sigma^2} = \log{(1 - c_h)} + \log{\theta_m} -\frac{(\logit(c^*) -  \mu_1)^2}{2\sigma^2}$$

$$\log{c_h} + \log{(1-\theta_m)} - \log{(1 - c_h)} + \log{\theta_m} = \frac{(\logit(c^*) -  \mu_0)^2}{2\sigma^2} -\frac{(\logit(c^*) -  \mu_1)^2}{2\sigma^2}$$

$$\logit(c_h) - \logit(\theta_m) = \frac{(\logit(c^*) -  \mu_0)^2 - (\logit(c^*) -  \mu_1)^2}{2\sigma^2}$$

$$2\sigma^2[\logit(c_h) - \logit(\theta_m)] = - 2 \logit(c^*) \mu_0 + 2 \logit(c^*) \mu_1 + \mu_0^2 - \mu_1^2$$

$$[2\sigma^2[\logit(c_h) - \logit(\theta_m)] + \mu_1^2 - \mu_0^2]/[2(\mu_1 - \mu_0)] = \logit(c^*)$$

$$\sigma^2[\logit(c_h) - \logit(\theta_m)]/(\mu_1 - \mu_0) + \frac{\mu_1 + \mu_0}2 = \logit(c^*) $$

$$c^* = \sigmoid\left(\frac{\sigma^2[\logit(c_h) - \logit(\theta_m)]}{\mu_1 - \mu_0} + \frac{\mu_1 + \mu_0}2\right)$$

\newpage
\section{Deriving $p(y_\text{m,h} = 1 \mid c_h )$}
\label{sec:appdxb}
$$p(y_\text{m,h} = 1 \mid c_h ) = \int_0^{c^*} c_h p(c_m) dc_m + \int_{c^*}^1 p(y_m=1|c_m) p(c_m) dc_m$$ 
$$= c_h\int_0^{c^*} \theta_m p(c_m | y = 1) + (1-\theta_m) p(c_m | y = 0) dc_m + \int_{c^*}^1 \theta_m p(c_m | y = 1) dc_m$$
$$= c_h\left[\theta_m P(c < c^* | y=1) + (1-\theta_m)P(c < c^* | y=0)\right] + \theta_m P(c^* < c | y = 1)$$

Since $$P(c < c^* | y=1)$$

$$= \Phi\left(\frac{\logit(c_m) - \mu_1}{\sigma} < \frac{\logit(c^*)- \mu_1}{\sigma}\right) $$

$$= \Phi\left(\frac{\sigma[\logit(c_h) - \logit(\theta_m)]}{\mu_1 - \mu_0} - \frac{\mu_1 - \mu_0}{2 \sigma}\right) $$

$$= \Phi\left(\frac{\logit(c_h) - \logit(\theta_m)}{\frac{\mu_1 - \mu_0}{\sigma}} - \frac12 \frac{\mu_1 - \mu_0}{\sigma}\right) $$

$$= \Phi\left(\frac{\logit(c_h) - \logit(\theta_m)}d - \frac d 2\right) $$

and $$P(c < c^* | y=0)$$

$$= \Phi\left(\frac{\logit(c_m) - \mu_0}{\sigma} < \frac{\logit(c^*)- \mu_0}{\sigma}\right) $$

$$= \Phi\left(\frac{\sigma[\logit(c_h) - \logit(\theta_m)]}{\mu_1 - \mu_0} + \frac{\mu_1 - \mu_0}{2 \sigma}\right) $$

$$= \Phi\left(\frac{\logit(c_h) - \logit(\theta_m)}{\frac{\mu_1 - \mu_0}{\sigma}} + \frac12 \frac{\mu_1 - \mu_0}{\sigma}\right) $$

$$= \Phi\left(\frac{\logit(c_h) - \logit(\theta_m)}d + \frac d 2\right) $$

Then $p(y_\text{m,h} = 1 \mid c_h )$

$$= c_h\left[\theta_m\Phi\left(\frac{\logit(c_h) - \logit(\theta_m)}d - \frac d 2\right) + (1-\theta_m)\Phi\left(\frac{\logit(c_h) - \logit(\theta_m)}d + \frac d 2\right)\right] $$$$ + \theta_m\left[1-\Phi\left(\frac{\logit(c_h) - \logit(\theta_m)}d - \frac d 2\right)\right]$$

$$= c_h\big[\theta_m\Phi(r) + (1-\theta_m)\Phi(r+d)\big] + \theta_m\left[1 - \Phi(r)\right]$$

where 

$$r = \frac{\logit(c_h) - \logit(\theta_m)}{d} - \frac{d}{2}$$

\newpage
\section{Deriving $p(y_\text{m,h} = 1 )$}
\label{sec:appdxc}

$$p(y_\text{m,h} = 1 )$$

$$=\int_0^1 p(c_h) p(y_\text{m,h} = 1 \mid c_h ) dc_h $$

$$= \int_0^1 p(c_h)c_h\left[\theta_m P(c < c^* | y=1) + (1-\theta_m)P(c < c^* | y=0)\right]dc_h + \theta_m\int_0^1 p(c_h)[1-P(c < c^* | y=1)] dc_h$$

$$=\int_0^1 p(c_h) c_h\left[\theta_m\Phi\left(\frac{\logit(c_h) - \logit(\theta_m)}d - \frac d 2\right) + (1-\theta_m)\Phi\left(\frac{\logit(c_h) - \logit(\theta_m)}d + \frac d 2\right)\right]dc_h $$$$+ \theta_m\int_0^1 p(c_h) \left[1-\Phi\left(\frac{\logit(c_h) - \logit(\theta_m)}d - \frac d 2\right)\right]dc_h$$

We derive the left and right integrals separately:

\subsection*{Right integral derivation} 

$$\theta_m\int_0^1 p(c_h)[1-P(c < c^* | y=1)] dc_h$$

$$=\theta_m\int_0^1 p(c_h) \left[1-\Phi\left(\frac{\logit(c_h) - \logit(\theta_m)}d - \frac d 2\right)\right]dc_h$$

$$=\theta_m\left[\int_0^1 p(c_h)dc_h - \int_0^1 p(c_h)\Phi\left(\frac{\logit(c_h) - \logit(\theta_m)}d - \frac d 2\right)dc_h\right]$$

$$=\theta_m\left[1 - \int_0^1 p(c_h)\Phi\left(\frac{\logit(c_h) - \logit(\theta_m)}d - \frac d 2\right)dc_h\right]$$

$$=\theta_m\left[1 - \int_0^1 \frac{\phi(\frac{\logit(c_h) - \mu_h}{\sigma_h})}{\sigma_h c_h (1-c_h)} \Phi\left(\frac{\logit(c_h) - \logit(\theta_m)}d - \frac d 2\right) dc_h\right]$$

Let $x = \frac{\logit(c_h) - \mu_h}{\sigma_h}$. Then

$$c_h = 0 \Rightarrow x = -\infty$$
$$c_h = 1 \Rightarrow x = \infty$$
$$dx = \frac1{\sigma_h c_h (1-c_h)} dc_h$$
$$\sigmoid(x \sigma_h + \mu_h) = c_h$$

and 

$$=\theta_m\left[1 - \int_{-\infty}^\infty \phi(x) \Phi\left(\frac{x \sigma_h + \mu_h - \logit(\theta_m)}d - \frac d 2\right) dx\right]$$

$$=\theta_m\left[1 - \int_{-\infty}^\infty \phi(x) \Phi\left(a+bx\right) dx\right]$$ 

$$=\theta_m\left[1 - \Phi\left(\frac{a}{\sqrt{1+b^2}}\right)\right]$$

where $a=\frac{\mu_h -\logit(\theta_m)}d - \frac d 2, b=\frac {\sigma_h} d$, based on Owen's table 10010.8 \citep{Owen1980-kl} $\int_{-\infty}^{+\infty} \phi(x) \Phi(a+bx) dx=\Phi\left(\frac{a}{\sqrt{1+b^2}}\right)$

\subsection*{Left integral derivation} 

$$\int_0^1 p(c_h)c_h\left[\theta_m P(c < c^* | y=1) + (1-\theta_m)P(c < c^* | y=0)\right]dc_h$$

$$=\int_0^1 p(c_h) c_h\left[\theta_m\Phi\left(\frac{\logit(c_h) - \logit(\theta_m)}d - \frac d 2\right) + (1-\theta_m)\Phi\left(\frac{\logit(c_h) - \logit(\theta_m)}d + \frac d 2\right)\right] $$

$$=\theta_m\int_0^1 p(c_h) c_h \Phi\left(\frac{\logit(c_h) - \logit(\theta_m)}d - \frac d 2\right) + (1-\theta_m)\int_0^1 p(c_h) c_h \Phi\left(\frac{\logit(c_h) - \logit(\theta_m)}d + \frac d 2\right) $$

For

$$\theta_m\int_0^1 p(c_h) c_h \Phi\left(\frac{\logit(c_h) - \logit(\theta_m)}d - \frac d 2\right)$$

Let $x = \frac{\logit(c_h) - \mu_h}{\sigma_h}$. Then

$$c_h = 0 \Rightarrow x = -\infty$$
$$c_h = 1 \Rightarrow x = \infty$$
$$dx = \frac1{\sigma_h c_h (1-c_h)} dc_h$$
$$\sigmoid(x \sigma_h + \mu_h) = c_h$$

and 
$$=\theta_m\int_{-\infty}^\infty \phi(x) \sigmoid(x \sigma_h + \mu_h) \Phi\left(\frac{x \sigma_h + \mu_h - \logit(\theta_m)}d - \frac d 2\right) dx$$

$$\approx \theta_m\int_{-\infty}^\infty \phi(x) \Phi(\lambda x \sigma_h + \lambda \mu_h) \Phi\left(\frac{x \sigma_h + \mu_h - \logit(\theta_m)}d - \frac d 2\right) dx$$ 

$$=\theta_m\int_{-\infty}^\infty \phi(x) \Phi(m+nx) \Phi\left(a+bx\right) dx$$ 

Similarly, 
$$(1-\theta_m)\int_0^1 p(c_h) c_h \Phi\left(\frac{\logit(c_h) - \logit(\theta_m)}d + \frac d 2\right)$$

$$\approx (1-\theta_m)\int_{-\infty}^\infty \phi(x) \Phi(m+nx) \Phi\left(a+d+bx\right) dx$$ 

Hence,

$$\approx \theta_m \bivariatenormal\left(\frac{a}{\sqrt{1+b^2}}, \frac{s}{\sqrt{1+t^2}}, \rho=\frac{b t}{\sqrt{1+b^2} \sqrt{1+t^2}}\right) $$$$ + (1-\theta_m) \bivariatenormal\left(\frac{a + d}{\sqrt{1+b^2}}, \frac{s}{\sqrt{1+t^2}}, \rho=\frac{b t}{\sqrt{1+b^2} \sqrt{1+t^2}}\right)$$

where $s=\lambda \mu_h, t=\lambda \sigma_h, a=\frac{\mu_h -\logit(\theta_m)}d - \frac d 2, b=\frac {\sigma_h} d$, based on Owen's table 20010.3 \citep{Owen1980-kl}: $$\int_{-\infty}^{+\infty} \Phi(a+bx) \Phi(m+nx) \phi(x) dx=\bivariatenormal\left(\frac{a}{\sqrt{1+b^2}}, \frac{s}{\sqrt{1+t^2}}, \rho=\frac{b t}{\sqrt{1+b^2} \sqrt{1+t^2}}\right)$$ and probit approximation in Murphy 10.5.2.2 \citep{Murphy2012-ek}: $$\sigmoid(a) \approx \Phi(\lambda x), \lambda=\sqrt{\frac{\pi}8}$$

\subsection*{Adding right and left integral}

Hence, the solution is 

\begin{align*}
& p(y_\text{m,h} = 1 ) =\int_0^1 p(c_h) p(y_\text{m,h} = 1 \mid c_h ) dc_h \\
&\approx \theta_m \bivariatenormal\left(\frac{a}{\sqrt{1+b^2}}, \frac{s}{\sqrt{1+t^2}}, \rho=\frac{b t}{\sqrt{1+b^2} \sqrt{1+t^2}}\right) \\
& + (1-\theta_m) \bivariatenormal\left(\frac{a + d}{\sqrt{1+b^2}}, \frac{s}{\sqrt{1+t^2}}, \rho=\frac{b t}{\sqrt{1+b^2} \sqrt{1+t^2}}\right) \\
& + \theta_m\left[1 - \Phi\left(\frac{a}{\sqrt{1+b^2}}\right)\right]
\end{align*}

 where
\begin{gather*}
a=\frac{\mu_h -\logit(\theta_m)}d - \frac d 2, \quad b=\frac {\sigma_h} d \\
s=\lambda \mu_h, \quad t=\lambda \sigma_h
\end{gather*}

\section{Dataset and Model Details} \label{app:addmd}
\subsection*{NLP dataset} The first real-world dataset is based on LLM benchmarks designed for uncertainty quantification \citep{Ye2024-iq}. In this study, the authors modified five representative NLP tasks: MMLU for question answering, CosmosQA for reading comprehension, HellaSwag for commonsense inference, HaluDial for dialogue response selection, and HaluSum for document summarization. Each task consists of 10,000 multiple-choice questions with six answer options, where the accuracy for random guessing is \(1/6\). The authors evaluated nine open-source LLM series: Llama-2, Mistral-7B, Falcon, MPT-7B, Gemma-7B, Qwen, Yi, DeepSeek, and InternLM-7B. Each series included different model variants based on model size, question prompt format, and instruction fine-tuning, as detailed in their paper. There are a total of 102 LLM variations for each benchmark.

We use the LLM logits based on token likelihoods for the multiple-choice options to compute accuracy and metacognitive sensitivity (AUC) for correct and incorrect answers for each model variant and task (note that the AUC we compute is distinct from the typical ROC AUC used to assess discrimination between positive and negative classes). Figure \ref{fig:auc_vs_acc} reveals a wide range of accuracy and AUC values for the CosmoQA benchmarks. Notably, AUC and accuracy are not perfectly correlated with Pearson correlation $R=0.84, p<0.001)$ (e.g., the three Llama variants in CosmosQA exhibit low accuracy but high AUC), making this dataset particularly suitable for testing the trade-offs between accuracy and metacognitive sensitivity.

\subsection*{Image Classification Dataset} 
The second dataset involves the ImageNet-16H image classification dataset introduced by \cite{Steyvers2022-xg}. The authors selected 1,200 images spanning 16 categories from the ImageNet challenge \citep{Deng2009-do}. To increase task difficulty, they applied phase noise distortion with spatial frequencies of 80, 95, 110, and 125. The authors also selected five classic image classifiers—AlexNet, DenseNet121, GoogLeNet, ResNet152, and VGG19—pretrained on ImageNet data and fine-tuned them on the noisy images at four different levels of epochs. This resulted in a total of 20 model variants (4 fine-tuning epochs \(\times\) 5 classifiers) per noise level. Figure \ref{fig:auc_vs_acc} shows the relationship between metacognitive sensitivity and model accuracy for a noise level of 125. The correlation is stronger than that observed for the NLP dataset, with \( R = 0.90, p < 0.001 \).

\bibliography{steyversbib,cogscipaperpileold,paperpile} 

\begin{thebibliography}{43}
\providecommand{\natexlab}[1]{#1}
\providecommand{\url}[1]{\texttt{#1}}
\expandafter\ifx\csname urlstyle\endcsname\relax
  \providecommand{\doi}[1]{doi: #1}\else
  \providecommand{\doi}{doi: \begingroup \urlstyle{rm}\Url}\fi

\bibitem[Bansal et~al.(2019)Bansal, Nushi, Kamar, Lasecki, Weld, and Horvitz]{Bansal2019-kl}
Gagan Bansal, Besmira Nushi, Ece Kamar, Walter~S Lasecki, Daniel~S Weld, and Eric Horvitz.
\newblock Beyond accuracy: The role of mental models in human-{AI} team performance.
\newblock \emph{Proceedings of the AAAI Conference on Human Computation and Crowdsourcing}, 7:\penalty0 2--11, October 2019.

\bibitem[Benjamin et~al.(2023)Benjamin, Morstatter, Abbas, Abeliuk, Atanasov, Bennett, Beger, Birari, Budescu, Catasta, et~al.]{benjamin2023hybrid}
Daniel~M Benjamin, Fred Morstatter, Ali~E Abbas, Andres Abeliuk, Pavel Atanasov, Stephen Bennett, Andreas Beger, Saurabh Birari, David~V Budescu, Michele Catasta, et~al.
\newblock Hybrid forecasting of geopolitical events.
\newblock \emph{AI Magazine}, 44\penalty0 (1):\penalty0 112--128, 2023.

\bibitem[Benz and Rodriguez(2023)]{Benz2023-vh}
Nina L~Corvelo Benz and Manuel~Gomez Rodriguez.
\newblock Human-aligned calibration for {AI}-assisted decision making.
\newblock \emph{arXiv [cs.LG]}, pages 14609--14636, May 2023.

\bibitem[Breiman(1996)]{Breiman1996-vc}
Leo Breiman.
\newblock Bagging predictors.
\newblock \emph{Mach. Learn.}, 24\penalty0 (2):\penalty0 123--140, August 1996.

\bibitem[Deng et~al.(2009)Deng, Dong, Socher, Li, Li, and Fei-Fei]{Deng2009-do}
Jia Deng, Wei Dong, Richard Socher, Li-Jia Li, Kai Li, and Li~Fei-Fei.
\newblock {ImageNet}: A large-scale hierarchical image database.
\newblock In \emph{2009 IEEE Conference on Computer Vision and Pattern Recognition}, pages 248--255. IEEE, June 2009.

\bibitem[Fleming and Lau(2014)]{Fleming2014-wr}
Stephen~M Fleming and Hakwan~C Lau.
\newblock How to measure metacognition.
\newblock \emph{Front. Hum. Neurosci.}, 8:\penalty0 443, July 2014.

\bibitem[Franklin and Frank(2020)]{Franklin2020-da}
Nicholas~T Franklin and Michael~J Frank.
\newblock Generalizing to generalize: Humans flexibly switch between compositional and conjunctive structures during reinforcement learning.
\newblock \emph{PLoS Comput. Biol.}, 16\penalty0 (4):\penalty0 e1007720, April 2020.

\bibitem[Gal and Ghahramani(2015)]{Gal2015-qk}
Y~Gal and Zoubin Ghahramani.
\newblock Dropout as a bayesian approximation: Representing model uncertainty in deep learning.
\newblock \emph{ICML}, 48:\penalty0 1050--1059, June 2015.

\bibitem[Galvin et~al.(2003)Galvin, Podd, Drga, and Whitmore]{Galvin2003-tb}
Susan~J Galvin, John~V Podd, Vit Drga, and John Whitmore.
\newblock Type 2 tasks in the theory of signal detectability: discrimination between correct and incorrect decisions.
\newblock \emph{Psychon. Bull. Rev.}, 10\penalty0 (4):\penalty0 843--876, December 2003.

\bibitem[Gaube et~al.(2025)Gaube, Jussupow, Kokje, Khan, Bondi-Kelly, Schicho, Kitamura, Koch, Ezer, Mottok, Lermer, Ghassemi, and Colak]{Gaube2025-fo}
Susanne Gaube, Ekaterina Jussupow, Eesha Kokje, Jowaria Khan, Elizabeth Bondi-Kelly, Andreas Schicho, Felipe~Campos Kitamura, Timo~Kevin Koch, Timur Ezer, Jürgen Mottok, Eva Lermer, Marzyeh Ghassemi, and Errol Colak.
\newblock Underreliance harms human-{AI} collaboration more than overreliance in medical imaging.
\newblock April 2025.

\bibitem[Goertzel(2014)]{Goertzel2014-cc}
Ben Goertzel.
\newblock Artificial general intelligence: Concept, state of the art, and future prospects.
\newblock \emph{J. Artif. Gen. Intell.}, 5\penalty0 (1):\penalty0 1--48, December 2014.

\bibitem[Grimaldi et~al.(2015)Grimaldi, Lau, and Basso]{Grimaldi2015-de}
Piercesare Grimaldi, Hakwan Lau, and Michele~A Basso.
\newblock There are things that we know that we know, and there are things that we do not know we do not know: Confidence in decision-making.
\newblock \emph{Neurosci. Biobehav. Rev.}, 55:\penalty0 88--97, August 2015.

\bibitem[{Imagining the Digital Future Center}(2025)]{Imagining-the-Digital-Future-Center2025-xb}
{Imagining the Digital Future Center}.
\newblock Close encounters of the {AI} kind: Main report.
\newblock \url{https://imaginingthedigitalfuture.org/reports-and-publications/close-encounters-of-the-ai-kind/close-encounters-of-the-ai-kind-main-report/}, March 2025.
\newblock Accessed: 2025-5-13.

\bibitem[Jiang et~al.(2020)Jiang, Araki, Ding, and Neubig]{Jiang2020-bv}
Zhengbao Jiang, Jun Araki, Haibo Ding, and Graham Neubig.
\newblock How can we know when language models know? on the calibration of language models for question answering.
\newblock \emph{arXiv [cs.CL]}, December 2020.

\bibitem[Kadavath et~al.(2022)Kadavath, Conerly, Askell, Henighan, Drain, Perez, Schiefer, Hatfield-Dodds, DasSarma, Tran-Johnson, Johnston, El-Showk, Jones, Elhage, Hume, Chen, Bai, Bowman, Fort, Ganguli, Hernandez, Jacobson, Kernion, Kravec, Lovitt, Ndousse, Olsson, Ringer, Amodei, Brown, Clark, Joseph, Mann, McCandlish, Olah, and Kaplan]{Kadavath2022-fj}
Saurav Kadavath, Tom Conerly, Amanda Askell, Tom Henighan, Dawn Drain, Ethan Perez, Nicholas Schiefer, Zac Hatfield-Dodds, Nova DasSarma, Eli Tran-Johnson, Scott Johnston, Sheer El-Showk, Andy Jones, Nelson Elhage, Tristan Hume, Anna Chen, Yuntao Bai, Sam Bowman, Stanislav Fort, Deep Ganguli, Danny Hernandez, Josh Jacobson, Jackson Kernion, Shauna Kravec, Liane Lovitt, Kamal Ndousse, Catherine Olsson, Sam Ringer, Dario Amodei, Tom Brown, Jack Clark, Nicholas Joseph, Ben Mann, Sam McCandlish, Chris Olah, and Jared Kaplan.
\newblock Language models (mostly) know what they know.
\newblock \emph{arXiv [cs.CL]}, July 2022.

\bibitem[Keswani et~al.(2021)Keswani, Lease, and Kenthapadi]{Keswani2021-vb}
Vijay Keswani, Matthew Lease, and Krishnaram Kenthapadi.
\newblock Towards unbiased and accurate deferral to multiple experts.
\newblock In \emph{Proceedings of the 2021 AAAI/ACM Conference on AI, Ethics, and Society}, New York, NY, USA, July 2021. ACM.

\bibitem[Kraemer(2014)]{Kraemer2014-or}
Helena~Chmura Kraemer.
\newblock \emph{Effect Size}, pages 1--3.
\newblock John Wiley \& Sons, Inc., Hoboken, NJ, USA, May 2014.

\bibitem[Lake et~al.(2011)Lake, Salakhutdinov, Gross, and Tenenbaum]{Lake2011-nm}
Brenden Lake, Ruslan Salakhutdinov, Jason Gross, and Joshua Tenenbaum.
\newblock One shot learning of simple visual concepts.
\newblock \emph{Proceedings of the Annual Meeting of the Cognitive Science Society}, 33\penalty0 (33), 2011.

\bibitem[Lee et~al.(2025)Lee, Pruitt, Zhou, Du, and Odegaard]{Lee2025-jm}
Doyeon Lee, Joseph Pruitt, Tianyu Zhou, Jing Du, and Brian Odegaard.
\newblock Metacognitive sensitivity: The key to calibrating trust and optimal decision-making with {AI}.
\newblock \emph{PNAS Nexus}, page gaf133, April 2025.

\bibitem[Lee et~al.(2015)Lee, O'Doherty, and Shimojo]{Lee2015-ms}
Sang~Wan Lee, John~P O'Doherty, and Shinsuke Shimojo.
\newblock Neural computations mediating one-shot learning in the human brain.
\newblock \emph{PLoS Biol.}, 13\penalty0 (4):\penalty0 e1002137, April 2015.

\bibitem[Liang et~al.(2022)Liang, Sloane, Donkin, and Newell]{liang2022adapting}
Garston Liang, Jennifer~F Sloane, Christopher Donkin, and Ben~R Newell.
\newblock Adapting to the algorithm: how accuracy comparisons promote the use of a decision aid.
\newblock \emph{Cognitive research: principles and implications}, 7\penalty0 (1):\penalty0 14, 2022.

\bibitem[Ma et~al.(2024)Ma, Wang, Lei, Shi, Yin, and Ma]{Ma2024-aq}
Shuai Ma, Xinru Wang, Ying Lei, Chuhan Shi, Ming Yin, and Xiaojuan Ma.
\newblock “are you really sure?” understanding the effects of human self-confidence calibration in {AI}-assisted decision making.
\newblock In \emph{Proceedings of the CHI Conference on Human Factors in Computing Systems}, volume~63, pages 1--20, New York, NY, USA, May 2024. ACM.

\bibitem[Madras et~al.(2017)Madras, Pitassi, and Zemel]{Madras2017-hw}
David Madras, T~Pitassi, and R~Zemel.
\newblock Predict responsibly: Improving fairness and accuracy by learning to defer.
\newblock \emph{Neural Inf Process Syst}, 31:\penalty0 6150--6160, November 2017.

\bibitem[Mandel and Barnes(2014)]{mandel2014accuracy}
David~R Mandel and Alan Barnes.
\newblock Accuracy of forecasts in strategic intelligence.
\newblock \emph{Proceedings of the National Academy of Sciences}, 111\penalty0 (30):\penalty0 10984--10989, 2014.

\bibitem[Maniscalco and Lau(2012)]{Maniscalco2012-xx}
Brian Maniscalco and Hakwan Lau.
\newblock A signal detection theoretic approach for estimating metacognitive sensitivity from confidence ratings.
\newblock \emph{Conscious. Cogn.}, 21\penalty0 (1):\penalty0 422--430, March 2012.

\bibitem[Murphy(1973)]{Murphy1973-ap}
Allan~H Murphy.
\newblock A new vector partition of the probability score.
\newblock \emph{J. Appl. Meteorol.}, 12\penalty0 (4):\penalty0 595--600, June 1973.

\bibitem[Murphy(2012)]{Murphy2012-ek}
Kevin~P Murphy.
\newblock \emph{Machine Learning: A Probabilistic Perspective}.
\newblock Adaptive Computation and Machine Learning series. MIT Press, London, England, August 2012.

\bibitem[Owen(1980)]{Owen1980-kl}
D~B Owen.
\newblock A table of normal integrals: A table.
\newblock \emph{Commun. Stat. Simul. Comput.}, 9\penalty0 (4):\penalty0 389--419, January 1980.

\bibitem[Peters(2022)]{Peters2022-xl}
Megan A~K Peters.
\newblock \emph{Confidence in decision-making}.
\newblock Oxford University Press, March 2022.

\bibitem[Platt(1999)]{Platt1999-bn}
J~Platt.
\newblock Probabilistic outputs for support vector machines and comparisons to regularized likelihood methods.
\newblock \emph{Advances in large margin classifiers}, 1999.

\bibitem[Rechkemmer and Yin(2022)]{Rechkemmer2022-ah}
Amy Rechkemmer and Ming Yin.
\newblock When confidence meets accuracy: Exploring the effects of multiple performance indicators on trust in machine learning models.
\newblock In \emph{CHI Conference on Human Factors in Computing Systems}, pages 1--14, New York, NY, USA, April 2022. ACM.

\bibitem[Siegert(2017)]{Siegert2017-fs}
Stefan Siegert.
\newblock Simplifying and generalising murphy's brier score decomposition: Simplification of brier score decomposition.
\newblock \emph{Q. J. R. Meteorol. Soc.}, 143\penalty0 (703):\penalty0 1178--1183, January 2017.

\bibitem[Steyvers and Kumar(2023)]{Steyvers2023-ve}
M~Steyvers and Aakriti Kumar.
\newblock Three challenges for {AI}-assisted decision-making.
\newblock \emph{Perspect. Psychol. Sci.}, 19\penalty0 (5):\penalty0 722--734, July 2023.

\bibitem[Steyvers et~al.(2014)Steyvers, Wallsten, Merkle, and Turner]{steyvers2014evaluating}
Mark Steyvers, Thomas~S Wallsten, Edgar~C Merkle, and Brandon~M Turner.
\newblock Evaluating probabilistic forecasts with bayesian signal detection models.
\newblock \emph{Risk Analysis}, 34\penalty0 (3):\penalty0 435--452, 2014.

\bibitem[Steyvers et~al.(2022)Steyvers, Tejeda, Kerrigan, and Smyth]{Steyvers2022-xg}
Mark Steyvers, Heliodoro Tejeda, Gavin Kerrigan, and Padhraic Smyth.
\newblock Bayesian modeling of human-{AI} complementarity.
\newblock \emph{Proc. Natl. Acad. Sci. U. S. A.}, 119\penalty0 (11):\penalty0 e2111547119, March 2022.

\bibitem[Steyvers et~al.(2025)Steyvers, Tejeda, Kumar, Belem, Karny, Hu, Mayer, and Smyth]{Steyvers2025-yc}
Mark Steyvers, Heliodoro Tejeda, Aakriti Kumar, Catarina Belem, Sheer Karny, Xinyue Hu, Lukas~W Mayer, and Padhraic Smyth.
\newblock What large language models know and what people think they know.
\newblock \emph{Nat. Mach. Intell.}, pages 1--11, January 2025.

\bibitem[Tejeda et~al.(2022)Tejeda, Kumar, Smyth, and Steyvers]{Tejeda2022-wv}
Heliodoro Tejeda, Aakriti Kumar, Padhraic Smyth, and M~Steyvers.
\newblock {AI}-assisted decision-making: A cognitive modeling approach to infer latent reliance strategies.
\newblock \emph{Comput. Brain Behav.}, 5\penalty0 (4):\penalty0 491--508, October 2022.

\bibitem[Vodrahalli et~al.(2022)Vodrahalli, Gerstenberg, and Zou]{Vodrahalli2022-wh}
Kailas Vodrahalli, Tobias Gerstenberg, and James Zou.
\newblock Uncalibrated models can improve human-{{AI}} collaboration.
\newblock \emph{arXiv [cs. AI]}, February 2022.

\bibitem[Wilson and Dawson(2023)]{Wilson2023-vi}
Steve Wilson and Ads Dawson.
\newblock {OWASP} top 10 for {LLM} applications version 1.1.
\newblock October 2023.

\bibitem[Wu et~al.(2024)Wu, Meder, and Schulz]{Wu2024-fo}
Charley~M Wu, Björn Meder, and Eric Schulz.
\newblock Unifying principles of generalization: Past, present, and future.
\newblock \emph{Annu. Rev. Psychol.}, October 2024.

\bibitem[Ye et~al.(2024)Ye, Yang, Pang, Wang, Wong, Yilmaz, Shi, and Tu]{Ye2024-iq}
Fanghua Ye, Mingming Yang, Jianhui Pang, Longyue Wang, Derek~F Wong, Emine Yilmaz, Shuming Shi, and Zhaopeng Tu.
\newblock Benchmarking {LLMs} via uncertainty quantification.
\newblock In \emph{The Thirty-eight Conference on Neural Information Processing Systems Datasets and Benchmarks Track}, 2024.

\bibitem[Zhang et~al.(2020{\natexlab{a}})Zhang, Liao, and Bellamy]{Zhang2020-gi}
Yunfeng Zhang, Q~Vera Liao, and Rachel K~E Bellamy.
\newblock Effect of confidence and explanation on accuracy and trust calibration in {AI}-assisted decision making.
\newblock In \emph{Proceedings of the 2020 Conference on Fairness, Accountability, and Transparency}, New York, NY, USA, January 2020{\natexlab{a}}. ACM.

\bibitem[Zhang et~al.(2020{\natexlab{b}})Zhang, Liao, and Bellamy]{Zhang2020-yf}
Yunfeng Zhang, Q~Vera Liao, and Rachel K~E Bellamy.
\newblock Effect of confidence and explanation on accuracy and trust calibration in {AI}-assisted decision making.
\newblock In \emph{Proceedings of the 2020 Conference on Fairness, Accountability, and Transparency}, New York, NY, USA, January 2020{\natexlab{b}}. ACM.

\end{thebibliography}

\end{document}